\documentclass[conference]{IEEEtran}
\IEEEoverridecommandlockouts
\usepackage{cite}
\usepackage{amsmath,amssymb,amsfonts}
\usepackage{algorithmic}
\usepackage{graphicx}
\usepackage{textcomp}
\usepackage{xcolor}
\usepackage{multicol}
\usepackage{multirow}
\usepackage{fancyhdr}
\def\BibTeX{{\rm B\kern-.05em{\sc i\kern-.025em b}\kern-.08em
    T\kern-.1667em\lower.7ex\hbox{E}\kern-.125emX}}
\begin{document}

\title{Leveraging Deep Learning with Multi-Head Attention for Accurate Extraction of Medicine from Handwritten Prescriptions}

\author
{\IEEEauthorblockN{Usman Ali}
\IEEEauthorblockA{\textit{Dept. of Computer Science} \\
\textit{GIFT University}\\
Gujranwala, Pakistan \\
usmanali@gift.edu.pk}
\and
\IEEEauthorblockN{Sahil Ranmbail}
\IEEEauthorblockA{\textit{Dept. of Computer Science} \\
\textit{GIFT University}\\
Gujranwala, Pakistan \\
201980059@gift.edu.pk}
\and
\IEEEauthorblockN{Muhammad Nadeem}
\IEEEauthorblockA{\textit{Dept. of Computer Science} \\
\textit{GIFT University}\\
Gujranwala, Pakistan \\
201980050@gift.edu.pk}
\and

\IEEEauthorblockN{Hamid Ishfaq}
\IEEEauthorblockA{\textit{Dept. of Computer Science} \\
\textit{GIFT University}\\
Gujranwala, Pakistan \\
201980038@gift.edu.pk}
\and
\hspace{3cm}\IEEEauthorblockN{Muhammad Umer Ramzan}
\IEEEauthorblockA{\hspace{4cm}\textit{Dept. of Computer Science} \\
\hspace{3.65cm}\textit{GIFT University}\\
\hspace{3.65cm}Gujranwala, Pakistan \\
\hspace{3.65cm}umer.ramzan@gift.edu.pk}
\and
\IEEEauthorblockN{Waqas Ali}
\IEEEauthorblockA{\textit{Dept. of Electrical Engineering (RCET)} \\
\textit{University of Engg. \& Tech. Lahore}\\
Lahore, Pakistan \\
waqas.ali@uet.edu.pk}

}
\IEEEoverridecommandlockouts
\IEEEpubid{\makebox[\columnwidth]{978-1-7281-3058-3/19/\$31.00 © 2024 IEEE \hfill}
\hspace{\columnsep}\makebox[\columnwidth]{ }}
\maketitle
 
\begin{abstract}
Extracting medication names from handwritten doctor prescriptions is challenging due to the wide variability in handwriting styles and prescription formats. This paper presents a robust method for extracting medicine names using a combination of Mask R-CNN and Transformer-based Optical Character Recognition (TrOCR) with Multi-Head Attention and Positional Embeddings. A novel dataset, featuring diverse handwritten prescriptions from various regions of Pakistan, was utilized to fine-tune the model on different handwriting styles. The Mask R-CNN model segments the prescription images to focus on the medicinal sections, while the TrOCR model, enhanced by Multi-Head Attention and Positional Embeddings, transcribes the isolated text. The transcribed text is then matched against a pre-existing database for accurate identification. The proposed approach achieved a character error rate (CER) of 1.4\% on standard benchmarks, highlighting its potential as a reliable and efficient tool for automating medicine name extraction.
\end{abstract}


\section{Introduction}
Handwritten doctor's prescriptions have long frustrated both patients and pharmacists. The difficulty in deciphering a doctor's handwriting goes beyond inconvenience; when the handwriting is unclear, it can prevent patients from accurately following medical advice. This problem is compounded by the variability in handwriting styles and prescription formats, making it challenging to develop a standardized text recognition method.

Although significant progress has been made, these methods still have certain limitations. Many existing approaches rely on Convolutional Neural Networks (CNNs) and Recurrent Neural Networks (RNNs), which, while effective, struggle with the variability and complexity of handwritten text \cite{18}, \cite{19}, \cite{20}. Techniques such as Bidirectional Long Short-Term Memory (BiLSTM) and Connectionist Temporal Classification (CTC) have also been used, but they often perform poorly on blurry or distorted images \cite{5}, \cite{3}. Furthermore, conventional models require extensive pre-processing and post-processing steps, which complicate the systematization of these methods \cite{2}, \cite{11}, \cite{9}.

To address these limitations, our research makes several significant contributions:

\begin{itemize}
    \item \textbf{Proposed a Hybrid Approach}: Utilized Mask R-CNN for segmentation and TrOCR for text recognition, followed by the implementation of string matching techniques to accurately retrieve medicine names from the recognized text, comparing results against an existing database.
    \item \textbf{Developed an Original Dataset}: Created a unique dataset from Pakistan containing various handwritten prescriptions, enabling the model to train on diverse handwriting styles and layouts.
\end{itemize}

\section{Literature Review}
In recent years, numerous researchers have explored various approaches to text recognition. Despite significant advancements in OCR technology, recognizing handwritten prescriptions from doctors and accurately identifying the medical information within them remains a substantial challenge. Handwriting styles vary greatly, as does the quality of the uploaded images. The existing literature on text recognition can generally be divided into two categories: broad context handwritten text extraction and the specific application of recognizing medicine names from handwritten prescriptions.

In \cite{3}, the authors employed a combination of CNN, BiLSTM, and Connectionist Temporal Classification (CTC) for text recognition. Although effective, the model encounters difficulties with blurred images. Additionally, the authors applied a CRNN model to the IAM dataset, leveraging both convolutional and recurrent layers, and achieved a CER of 4.57\%, demonstrating the model's effectiveness in recognizing distorted text sequences. The incorporation of CTC further enhances the model's ability to process unsegmented data, making it highly configurable and accurate across various text recognition tasks. 
In \cite{16}, the authors proposed a BCHWTR system specifically designed for Indian bank cheques. The system incorporates various techniques, including Histogram of Oriented Gradients feature decomposition, region masking of segmented images, and texture extraction using a gray-level co-occurrence matrix, followed by classification with a Support Vector Machine (SVM). This approach effectively addresses the challenges posed by variations in handwriting and complex backgrounds on cheque fields.
In \cite{15}, the authors utilized a methodology where preprocessed images are first input into a 5-layer CNN for feature extraction, followed by a two-layer RNN with LSTM units. However, this technique struggles with noisy or broken text.
In \cite{26}, the authors describe a system that employs OCR technology, specifically using Tesseract, to capture images via an Android device's camera. The system involves pre-processing techniques before segmentation, followed by feature extraction to identify handwritten characters and convert them into a digitally editable format through post-processing steps. A significant limitation of the system is that character recognition can be computationally intensive and slow when dealing with various handwriting styles, and the accuracy largely depends on the quality of the handwritten data.
Moreover, the system described in \cite{1} utilizes a Convolutional Recurrent Neural Network (CRNN), integrating CNNs and Recurrent Neural Networks (RNNs) for the recognition of handwritten English medical prescriptions. This approach involves image segmentation, character detection, and classification into 64 predefined characters using a Deep Convolution Recurrent Neural Network. A significant limitation of this study is its dependence on the IAM and self-generated datasets, rather than using images directly collected from doctors. This could result in misleading outcomes when the system is applied to real-world prescriptions from medical professionals.
In \cite{22}, a CNN-based system is utilized to recognize handwritten medicine names from doctors' prescriptions. The process includes preprocessing, feature extraction, and post-processing with OCR, resulting in a Character Error Rate (CER) of 8.8\% on the IAM dataset, which is inadequate for practical applications. Another study in \cite{34}, introduced a system employing a Deep Convolutional Recurrent Neural Network (CRNN) to recognize text in doctors' cursive handwriting, achieving a training accuracy of 76\% and a validation accuracy of 72\% on a dataset of 540 prescription images.
In \cite{10}, a Neural Network model is employed to recognize text in doctors' handwritten prescriptions. The approach utilizes the EMNIST dataset, with images preprocessed to grayscale, segmented into individual characters, and recognized using a three-layer Neural Network. The model achieved a training accuracy of 94.49\% and a test accuracy of 80.72\%. However, a notable limitation of this study is the confusion between characters that appear similar.
In \cite{0}, the methodology involves preprocessing using techniques such as stroke enhancement, followed by training with LayoutLMv2 and DONUT, and utilizing classifiers like Gradient Boosting and Random Forest for classifying medicine types. This approach was tested on 200 real and 300 synthetic images; however, the classifiers demonstrated inconsistent performance when compared to more established models like LayoutLMv2 and DONUT.
The paper titled \cite{trocr} "TrOCR: Transformer-based Optical Character Recognition with Pre-trained Models" by Minghao Li and colleagues introduces TrOCR, a groundbreaking OCR model that utilizes Transformer architectures for both image interpretation and text generation. Unlike traditional approaches that depend on CNNs for visual feature extraction and RNNs for text decoding, TrOCR integrates pre-trained image Transformers, such as DeiT and BEiT, with text Transformers like RoBERTa and MiniLM in a unified framework. This integration streamlines the model by eliminating the need for convolutional networks and external language models. TrOCR sets new benchmarks in OCR tasks, including printed, handwritten, and scene text recognition, surpassing existing models without requiring complex pre- or post-processing steps. The model is pre-trained on large-scale synthetic data and fine-tuned on human-annotated datasets, demonstrating its effectiveness and versatility across various OCR applications.
In \cite{maskrcnn}, Detectron, developed by Facebook AI Research (FAIR), is presented as an open-source framework that implements advanced object detection algorithms, including Mask R-CNN. This versatile and robust framework is designed for research and production environments, supporting tasks such as instance segmentation and keypoint detection. Mask R-CNN, a key component of Detectron, builds upon Faster R-CNN by adding a branch for predicting segmentation masks, delivering state-of-the-art performance on benchmarks like COCO. Additionally, it introduces RoIAlign to resolve misalignment issues, significantly improving mask accuracy.
\section{Dataset and Description}
In this section, we will describe the datasets used in this research, specifically focusing on the dataset of handwritten doctor prescriptions
\subsection{Image Dataset}
\begin{figure}[http]
    \centering
    \includegraphics[width=0.35\paperwidth, height=0.26\paperheight]{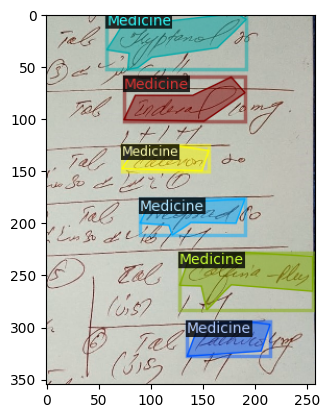}
    \caption{Annotated Image}
    \label{fig:segmentation2}
\end{figure}

To address the significant variability in handwriting styles and prescription formats, we conducted a comprehensive data collection effort, gathering approximately 1,000 handwritten prescriptions from 50 doctors across various regions of Pakistan, all with informed consent from both doctors and patients. This diverse and representative dataset ensures that our model is exposed to a wide spectrum of handwriting styles and formats, thereby enhancing its robustness and generalizability. The segmented region of the medicine, which is critical for subsequent processing, is illustrated in Figure~\ref{fig:segmentation2}. To enhance the robustness of our model, we applied data augmentation techniques, including brightness adjustment, contrast normalization, translation, minor shearing, elastic transformation, Gaussian noise, and cropping with padding. These augmentations expanded our dataset to a total of 9,920 images. 

\section{Proposed Methodology}\label{sec4}
\begin{figure*}[t]
    \centering
    \includegraphics[width=\textwidth]{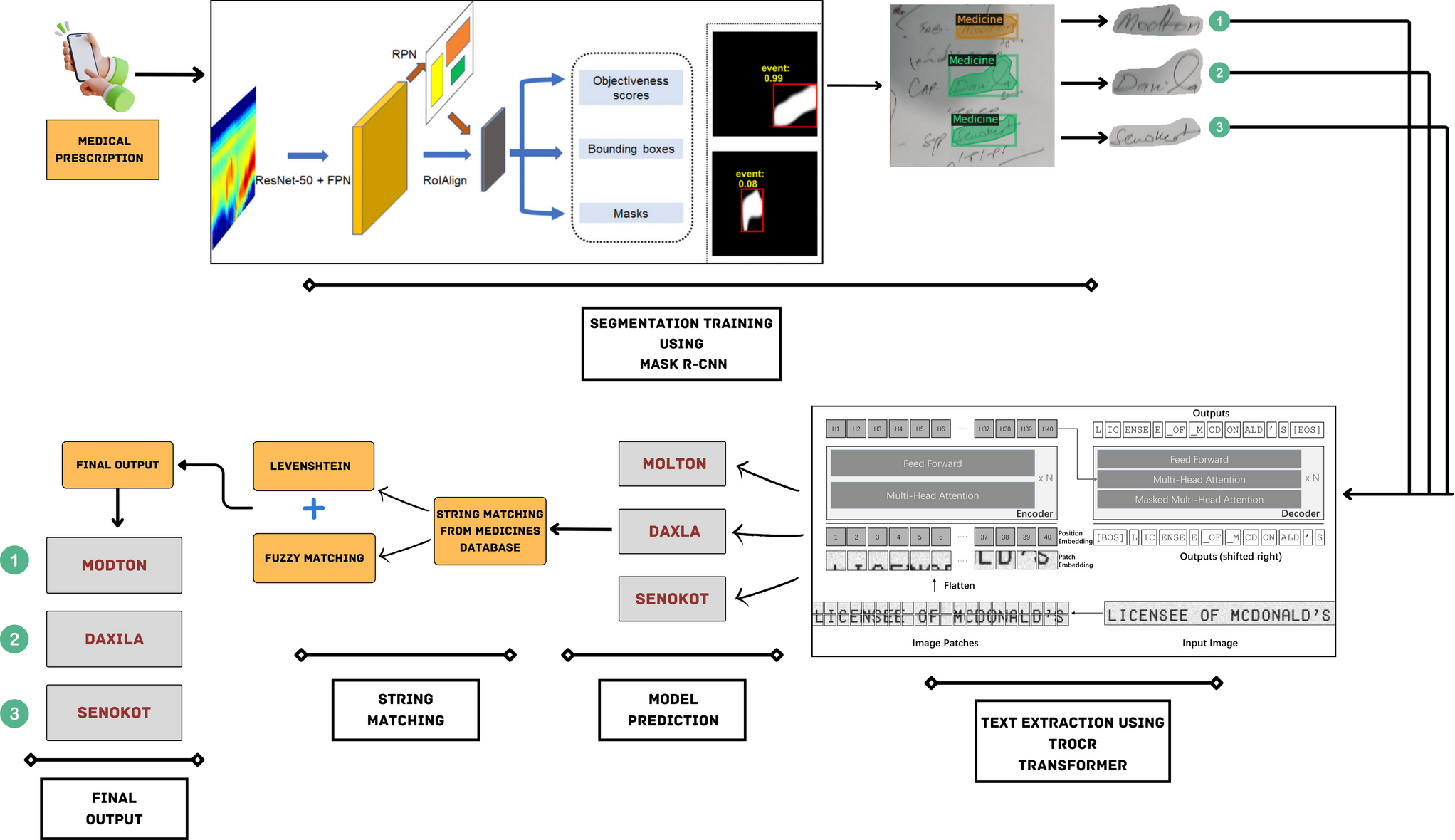} 
    \caption{Overall Workflow}
    \label{fig:overallWorkflow}
\end{figure*}

We propose an enhanced methodology to accurately extract and recognize handwritten medicine names from doctors’ prescriptions, shown in Fig~\ref{fig:overallWorkflow}. The process involves feature extraction using ResNet-50 with a Feature Pyramid Network (FPN), region identification via a Region Proposal Network (RPN), precise alignment with RoI Align, and text isolation through masking. Segmented regions are then processed by the TrOCR model for handwritten text recognition, followed by hybrid string matching for accurate medicine name identification.

The first phase of our methodology utilizes ResNet-50, a pre-trained CNN from ImageNet, which utilizes residual blocks with identity mapping \( y = F(x, \{W_i\}) + x \) to effectively preserve input signal integrity while learning complex features. To enhance the model's ability to detect objects at various scales, we integrate a FPN with ResNet-50. The FPN constructs a pyramid of features at multiple spatial scales, combining low-resolution, high-semantic information with high-resolution, low-semantic data. This fusion is crucial for detecting objects of different sizes within the prescriptions. Formally, the feature maps at each level \( l \) of the pyramid are computed as follows:
\begin{equation}
P_l = W_l \ast C_l + \text{upsample}(P_{l+1})
\end{equation}

where \( W_l \) denotes the convolutional weights applied at level \( l \), \( C_l \) represents the feature map from the ResNet layer, and \( \ast \) indicates convolution. 
Next, the Region Proposal Network (RPN) is employed to generate candidate regions, or anchor boxes, that likely contain medicine names. As a fully convolutional network, the RPN scans the feature maps produced by the FPN to propose regions of interest. For each anchor, the RPN outputs two key predictions: an objectness score, which indicates the probability of an object being present, and the bounding box coordinates that define the location of the object within the region. The RPN architecture includes a convolutional layer followed by two sibling layers, which are mathematically represented as:
\begin{equation}
\text{score} = \sigma(W_s \ast x + b_s)
\end{equation}

\begin{equation}
\text{bbox} = W_r \ast x + b_r
\end{equation}

where \( W_s \) and \( W_r \) are the weights, \( b_s \) and \( b_r \) are biases, and \( \sigma \) is the sigmoid activation function used for the objectness score. To ensure accurate feature extraction from the proposed regions, the RoI Align operation is applied. RoI Align refines the bounding boxes proposed by the RPN by mapping them onto a fixed-size feature map extracted from the FPN. This operation addresses the quantization errors inherent in traditional RoI Pooling through bilinear interpolation, which calculates the value at a specific point as:

\begin{equation}
v = \sum_{i=0}^{1}\sum_{j=0}^{1} w_{ij} \cdot x(p_i, p_j)
\end{equation}

where \( w_{ij} \) represents the weights based on the distance to the sample points, and \( x(p_i, p_j) \) are the sampled values. This process ensures smooth and precise alignment of features, crucial for accurate text recognition. The RoI alignment operation outputs: objectness scores, bounding boxes, and binary masks for each identified medicine name. The objectness scores reflect the likelihood that a region contains a medicine name, while the bounding boxes define the coordinates of the rectangular regions where these names are located. The mask prediction, essential for instance segmentation tasks, produces a binary map indicating the pixels corresponding to each medicine name within the bounding box. The mathematical formulation for these outputs is as follows:

\begin{equation}
\text{Objectness score: } s = \sigma(W_o \ast F + b_o)
\end{equation}

\begin{equation}
\text{Bounding box coordinates: } b = W_b \ast F + b_b
\end{equation}

\begin{equation}
\text{Mask prediction: } M = \sigma(W_m \ast F + b_m)
\end{equation}

\begin{figure}[h]
    \centering
    \includegraphics[width=0.5\textwidth]{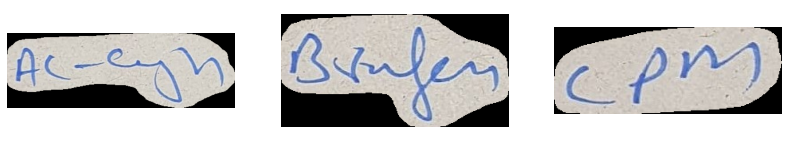}
    \caption{Segmentation of medicinal areas using Mask R-CNN}
    \label{fig:segmentation}
\end{figure}

Here, \( W_o \), \( W_b \), and \( W_m \) represent the weights for objectness, bounding box, and mask predictions, respectively, \( F \) denotes the feature maps and $\sigma$ denotes the sigmoid activation function. The model ultimately produces pixel-level binary masks for each detected medicine name, effectively distinguishing the text from the background and other prescription elements shown in Fig~\ref{fig:segmentation}.

In the second phase of our methodology, the segmented images of prescription areas obtained in the first phase are processed through the TrOCR model, which is specifically designed for handwritten text recognition \cite{trocr}. The process begins with an input image \( X \in \mathbb{R}^{H \times W \times C} \), where \( H \), \( W \), and \( C \) denote the height, width, and number of channels of the segmented prescription image. This image is divided into a grid of fixed-size patches \( P_i \in \mathbb{R}^{p \times p \times C} \), where \( p \) is the patch size. Each patch, representing a specific region of the handwritten text, is then flattened into a one-dimensional vector \( x_i \in \mathbb{R}^{p^2 \times C} \) as follows:

\begin{equation}
x_i = \text{Flatten}(P_i)
\end{equation}

To maintain the spatial structure of the prescription text, positional embeddings \( E_{\text{pos}}(i) \in \mathbb{R}^{d} \) are added to the patch embeddings, where \( d \) is the embedding dimension. This embedding retains information about the relative positions of patches within the original image, which is crucial for accurately interpreting the structure of handwritten text. The initial embedding for the \( i^{th} \) patch is computed as:

\begin{equation}
z_i^0 = x_i W_e + E_{\text{pos}}(i)
\end{equation}

Here, \( W_e \in \mathbb{R}^{p^2 \times d} \) is a learnable weight matrix used to project the flattened patch \( x_i \) into the embedding space. The embedded patches, now carrying both the content and positional information of the text segments, are passed through the Encoder and Decoder, based on the Vision Transformer (ViT) architecture. The Encoder utilizes a multi-head attention mechanism to capture relationships and features at multiple scales across the patches, which is essential for recognizing the varying sizes and orientations of handwritten characters. For each layer \( l \) in the Encoder, the multi-head attention output is computed as:

\begin{equation}
\text{Attention}(Q, K, V) = \text{softmax}\left(\frac{QK^T}{\sqrt{d_k}}\right)V
\end{equation}

where \( Q = z_i^{l-1}W_Q \), \( K = z_i^{l-1}W_K \), and \( V = z_i^{l-1}W_V \) are the query, key, and value matrices, respectively. \( W_Q \), \( W_K \), and \( W_V \) are learnable weight matrices, and \( d_k \) is the dimensionality of the key vectors. The attention output is then passed through a feed-forward neural network (FFNN) for further refinement:

\begin{equation}
z_i^l = \text{FFNN}(\text{Attention}(Q, K, V))
\end{equation}

This process is repeated across \( N \) layers, progressively refining the feature representations of the image patches, enabling the model to recognize intricate handwriting patterns. Following the Encoder, the Decoder, also based on the ViT architecture, applies another series of multi-head attention mechanisms. In this stage, the Decoder attends to various positions in the patch space, capturing intricate patterns and dependencies within the handwritten text patches. The attention output in the Decoder is computed similarly:

\begin{equation}
\text{Decoder Attention}(Q', K', V') = \text{softmax}\left(\frac{Q'K'^T}{\sqrt{d_k}}\right)V'
\end{equation}

where \( Q' = z_i^{\text{enc}}W'_Q \), \( K' = z_i^{\text{enc}}W'_K \), and \( V' = z_i^{\text{enc}}W'_V \), with \( z_i^{\text{enc}} \) representing the output from the Encoder. This output is further processed through a feed-forward network:

\begin{equation}
z_i^{\text{dec}} = \text{FFNN}(\text{Decoder Attention}(Q', K', V'))
\end{equation}

This procedure is repeated across \( N \) layers, with each layer enhancing the features detected from the handwritten text patches. Finally, the output of the TrOCR model is a sequence of text tokens, corresponding to the recognized medicine names from the prescription segments. These tokens, representing the extracted text, are generated as:

\begin{equation}
\text{Output Sequence} = \text{softmax}(W_o \cdot z_i^{\text{dec}} + b_o)
\end{equation}

where \( W_o \in \mathbb{R}^{d \times V} \) is a learnable weight matrix, \( b_o \in \mathbb{R}^{V} \) is a bias vector, and \( V \) is the vocabulary size of the text tokens. The softmax function ensures that the output probabilities sum to 1, effectively generating a sequence of recognized text tokens, representing the medicine names within the prescription images.
Lastly, the predicted medicine names are matched against the medicine column in our database using a hybrid string matching technique. This technique involves two main components: Levenshtein distance and fuzzy matching.

The Levenshtein distance \( D(w_1, w_2) \) between two words \( w_1 \) and \( w_2 \) is computed as the minimum number of single-character edits (insertions, deletions, or substitutions) required to change one word into the other. The similarity score \( S_{L}(w_1, w_2) \) is then calculated as:

\begin{equation}
S_{L}(w_1, w_2) = \left( 1 - \frac{D(w_1, w_2)}{\max(|w_1|, |w_2|)} \right) \times 100
\end{equation}

where \( |w_1| \) and \( |w_2| \) are the lengths of the words \( w_1 \) and \( w_2 \). If the similarity score \( S_{L}(w_1, w_2) \) exceeds a predefined threshold \( T_L \), the word \( w_2 \) is selected as a match:

\begin{equation}
\text{if } S_{L}(w_1, w_2) \geq T_L, \text{ then } w_2 \text{ is a match}
\end{equation}

When Levenshtein matching does not yield a result or requires additional validation, the fuzzy matching technique is employed. The similarity score \( S_{F}(w_1, w_2) \) is computed using the fuzzy string matching algorithm:

\begin{equation}
S_{F}(w_1, w_2) = \text{fuzz.ratio}(w_1, w_2)
\end{equation}

If the similarity score \( S_{F}(w_1, w_2) \) meets or exceeds a predefined threshold \( T_F \), the word \( w_2 \) is considered a potential match. The best match is selected based on the highest similarity score:

\begin{equation}
\text{Best match} = \max_{w_2 \in \text{lexicon}}(S_{F}(w_1, w_2))
\end{equation}

The final decision process can be described as:

\begin{equation}
\text{Output} = 
\begin{cases} 
w_2 & \text{if } S_{L}(w_1, w_2) \geq T_L \\ 
& S_{F}(w_1, w_2) \geq T_F \\ 
\text{"no"} & \text{if no match is found}
\end{cases}
\end{equation}

\section{Results and Discussion}\label{sec5}
In this section, we present the evaluation results of our Mask R-CNN and TrOCR Base Handwritten models and discuss the implications of these findings.

\subsection{Mask R-CNN Model}

The Mask R-CNN model segments medicinal areas from handwritten prescriptions. Performance metrics are summarized in Table \ref{tab:mask_rcnn_results}, which clearly presents the model's results for bounding box and segmentation tasks. 
\begin{table}[h]
    \centering
    \caption{Summary of Key Metrics for Mask R-CNN Model}
    \resizebox{0.3\textwidth}{!}{
    \begin{tabular}{|l|c|c|}
        \hline
        \textbf{Metric} & \textbf{Bbox} & \textbf{Segm} \\
        \hline
        AP & 51.251 & 48.299 \\
        AP50 & 77.227 & 79.354 \\
        AP75 & 60.046 & 62.434 \\
        APm & 54.201 & 49.172 \\
        APl & 48.096 & 47.708 \\
        \hline
    \end{tabular}
    }
    
    \label{tab:mask_rcnn_results}
\end{table}
\begin{table*}[t]
\centering
\caption{CER Comparison Across Models and Data Categories}
\label{tab:cer_comparison}
\renewcommand{\arraystretch}{1.3} 
\fontsize{10}{10}\selectfont 

\begin{tabular}{lcccc}
\hline
\textbf{Model} & \textbf{Data Categories} & \textbf{CER Before SM} & \textbf{CER After SM} & \textbf{Improvement} \\ \hline

\multirow{3}{*}{TrOCR-Base Printed} 
& Valid Data & 0.0127 & 0.0127 & - \\ 
& Pattern Change & 0.2007 & 0.1801 & \textbf{0.0206} \\ 
& Test & 0.2910 & 0.2937 & -0.0027 \\ \hline

\multirow{3}{*}{TrOCR-Base STR} 
& Valid Data & 0.0100 & 0.0105 & -0.0005 \\ 
& Pattern Change & 0.1202 & 0.0801 & \textbf{0.0401} \\ 
& Test & 0.1470 & 0.1421 & \textbf{0.0049} \\ \hline

\multirow{3}{*}{\textbf{TrOCR-Base Handwritten}} 
& \textbf{Valid Data} & 0.0143 & \textbf{0.0137} & \textbf{0.0006} \\ 
& \textbf{Pattern Change} & 0.0670 & \textbf{0.0386} & \textbf{0.0386} \\ 
& \textbf{Test} & 0.1544 & \textbf{0.1351} & \textbf{0.0193} \\ \hline

\end{tabular}

\end{table*}

\subsection{TrOCR Models}
In this experimentation, we evaluated the performance of several pretrained models on handwriting recognition tasks, with the primary focus on the TrOCR-Base model. 
We performed test experiments on these model using three different scenarios: \textbf{Valid Set:} The augmented dataset divided into a training set consisting of 7,928 images (80\%) and a Valid set containing 1,992 images (20\%).
 \textbf{Pattern Change:} Medical labels exist in the training set) This set includes images with new patterns but familiar medical labels, consisting of 67 new images.
\textbf{Test Set:} (No labels or patterns during training) This set contains completely new images with no labels or patterns seen during training, consisting of 47 images.

TrOCR-Base, a transformer-based OCR model, is specifically designed for text recognition, leveraging the powerful transformer architecture to tackle the complexities inherent in handwritten text, including varying handwriting styles and inconsistencies. This model's ability to manage diverse text forms makes it a robust choice for our study.

Among the pretrained models utilized, the TrOCR-Base-Handwritten model stands out as it is pretrained on handwritten datasets, making it particularly adept at recognizing cursive and other handwritten text forms. This model further fine-tuned on augmented handwritten data to enhance its robustness and accuracy. Additionally, we explored the TrOCR-Base-STR model, originally trained on Scene Text Recognition (STR) datasets containing text in natural scenes such as street signs and billboards. Although not initially intended for handwritten text, the STR model's robust text recognition capabilities warranted its inclusion in our study. Lastly, the TrOCR-Base-Printed model, trained on printed text data from books, documents, and other typeset materials, has also been evaluated on handwritten datasets to assess its limitations and potential for transfer learning in handwriting recognition tasks.

Based on the results shown in Table \ref{tab:cer_comparison}, the TrOCR-Base-Handwritten model is identified as the best-performing model for handwriting recognition. Its superior ability to handle variations in handwriting styles, as reflected in the significant CER reductions across all data categories, makes it the optimal choice for further experimentation and deployment.

\section{Conclusion}\label{sec6}
This paper introduces a robust method for extracting medication names from handwritten prescriptions, effectively addressing the challenges of varied handwriting styles and formats. Utilizing a combination of Mask R-CNN for segmentation and TrOCR for text recognition, our approach achieved a CER of 1.4\% on standard benchmarks. The use of a diverse dataset of handwritten prescriptions from Pakistan highlights the model's reliability and efficiency, making it a promising tool for automating medicine name extraction.

\bibliography{Ref}
\bibliographystyle{IEEEtran}

\end{document}